\documentclass[Afour,sageh,times]{sagej}

\usepackage{moreverb,url}
\usepackage{amsmath} 
\usepackage{amssymb}  
\usepackage{subfigure}
\usepackage{multirow}
\usepackage{array,booktabs}
\usepackage{diagbox}
\usepackage{balance}
\usepackage{mathtools}
\usepackage{subfigure}
\usepackage{verbatim}
\usepackage{adjustbox}
\usepackage{algorithm}
\usepackage{algpseudocode}
\usepackage{romannum}
\usepackage{amssymb}
\usepackage{pifont}
\usepackage{hyperref}
\usepackage{enumitem}
\hypersetup{
  colorlinks=true, 
  urlcolor=blue,   
  linkcolor=blue,  
  citecolor=black  
}
\usepackage[table]{xcolor}
\usepackage{rpm_math}
\usepackage{rpm_misc}
\usepackage{rpm_acronyms}
\usepackage{rpm_SIunits}
\usepackage{soul,color}
\usepackage{multirow}
\usepackage{soul,color}
\usepackage{lipsum}

\newcommand{\B}[1]{{\textbf{#1}}}

\newcommand\BibTeX{{\rmfamily B\kern-.05em \textsc{i\kern-.025em b}\kern-.08em
T\kern-.1667em\lower.7ex\hbox{E}\kern-.125emX}}

\def\volumeyear{2023}

\begin{document}

\runninghead{Kim \emph{et al.}}

\title{TRansPose: Large-Scale Multispectral Dataset for Transparent Object}

\author{Jeongyun Kim\affilnum{1$^\dagger$}, Myung-Hwan Jeon\affilnum{2$^\dagger$}, Sangwoo Jung\affilnum{1}, Wooseong Yang\affilnum{1}, Minwoo Jung\affilnum{1}, Jaeho Shin\affilnum{1}, Ayoung Kim\affilnum{1}}

\affiliation{\affilnum{$\dagger$}The authors contributed equally to this paper\\
\affilnum{1}Dept. of Mechanical Engineering, SNU, Seoul, S. Korea\\
\affilnum{2}Institute of Advanced Machines and Design, SNU, Seoul, S. Korea}

\corrauth{Ayoung Kim, Dept. of Mechanical Engineering, SNU, Seoul, S. Korea}
\email{ayoungk@snu.ac.kr}

\begin{abstract}
  
Transparent objects are encountered frequently in our daily lives, yet recognizing them poses challenges for conventional vision sensors due to their unique material properties, not being well perceived from RGB or depth cameras. Overcoming this limitation, thermal infrared cameras have emerged as a solution, offering improved visibility and shape information for transparent objects. In this paper, we present TRansPose, the first large-scale multispectral dataset that combines stereo RGB-D, \acf{TIR} images, and object poses to promote transparent object research. The dataset includes 99 transparent objects, encompassing 43 household items, 27 recyclable trashes, 29 chemical laboratory equivalents, and 12 non-transparent objects. It comprises a vast collection of 333,819 images and 4,000,056 annotations, providing instance-level segmentation masks, ground-truth poses, and completed depth information. The data was acquired using a FLIR A65 \acf{TIR} camera, two Intel RealSense L515 RGB-D cameras, and a Franka Emika Panda robot manipulator. Spanning 87 sequences, TRansPose covers various challenging real-life scenarios, including objects filled with water, diverse lighting conditions, heavy clutter, non-transparent or translucent containers, objects in plastic bags, and multi-stacked objects. Supplementary material can be accessed from the following link: \url{https://sites.google.com/view/transpose-dataset}
  
\end{abstract}

\keywords{Dataset, Multispectral Cameras, Object Recognition, Transparent Object, Object Pose Estimation}

\maketitle

\section{Introduction}

Over the past decade, object recognition has garnered significant attention, with datasets \citep{kasper-2012-kit, calli-2017-ycb, novkovic-2019-clubs} contributing greatly to advancements in the robotics research \citep{Saxena-2008-novel, sinapov-2011-interactive,  li-2011-segmentation, zeng-2022-pick}. These datasets, however, mainly provide RGB images and capture only a limited representation of transparent objects.
Transparent objects are common in daily life, recycling centers, as well as chemical and household settings, raising the requirement to perceive these objects for robots and automation. When recognizing transparent objects with vision sensors, they show insurmountable challenges caused by their inherent material. The transparent objects are absent of distinct features and textures caused by high dependency on the background, making inconsistent visual appearances. In addition, the usage of depth measurement obtained from the distance sensors is arduous since the transparent objects break the Lambertian reflectance.

Many existing methods have mainly targeted RGB-based approaches, proposing datasets for boosting the research of transparent object recognition, such as segmentation, detection, and pose estimation \citep{chen-2018-cvpr, liu-2020-cvpr, xie-2020-eccv, sajjan-2020-icra,  liu-2021-iccv, chen-2022-eccv, fang-2022-ral}. Still, these RGB-based datasets suffer from the inconsistent texture of transparent objects, which are susceptible to background interference. To tackle these challenges, alternative sensors have been examined, including light-field  \citep{xu-2015-iccv, zhou-2020-ral} and polarized cameras \citep{mei-2022-cvpr}. Despite their potential, these light-dependent sensors may lead to longer processing times or higher noise levels due to optical effects.

\begin{figure*}[!t]
    \centering
    \includegraphics[width=0.95\textwidth]{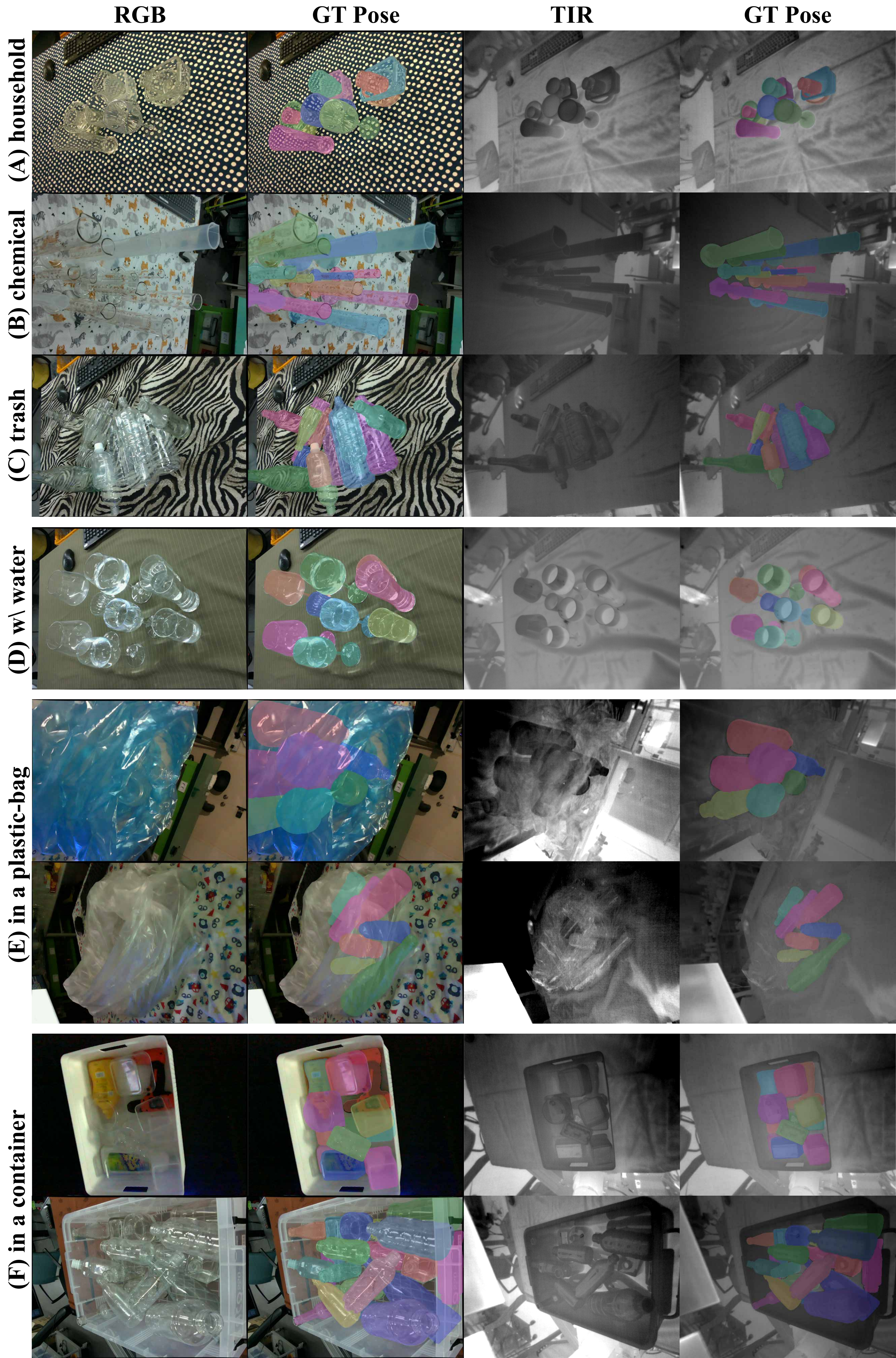}
    \caption{Sample RGB images and \ac{TIR} images with  mask images rendered using the provided annotations.}
    \label{fig:result}
\end{figure*}

Recently, another alternative sensor, \acf{TIR} camera, has been introduced in RGB-T dataset \citep{huo-2023-tip} for their robust sensing capability on transparent objects (specifically glasses). \ac{TIR} cameras function by measuring the temperature emitted from the object surface and operate within the wavelength range of 8-12 $\um$, which does not penetrate the typical material of transparent objects. Valuing a similar potential of \ac{TIR} but not being limited to glasses, our dataset aims to expand the potential of the \ac{TIR} imaging for a broad range of transparent objects. As illustrated in the \figref{fig:result}, the distinctive features of \ac{TIR} cameras facilitate a straightforward perception of the overall shape of transparent objects, thereby simplifying tasks such as segmentation and pose estimation.  Moreover, TIR cameras prove useful in real-life scenarios as they can penetrate vinyl, enabling observation of objects enclosed in plastic bags (\figref{fig:result}). Our dataset encompasses such instances captured to support further research in using \ac{TIR} for transparent objects.


\begin{table*}[!t]
\centering
\caption{Comparison with existing transparents datasets on sensor modality, the number of transparents objects (\textbf{\#transparent obj}) and total object (\textbf{\#total obj}), the number of frames in the real-world (\textbf{\#frame}), the total number of pose annotations (\textbf{\#pose annotation}), the total number of sequences (\textbf{\#seq}) and \textbf{object classes}. The mark \ding{55} indicates that dataset consist of unspecified number of objects like glass walls or window.}
\begin{adjustbox}{width=0.91\linewidth}
{%
\begin{tabular}{llcccccl}
\hline
\textbf{Dataset} & \textbf{Modality}      & \textbf{\#transparent obj} & \textbf{\#total obj} & \textbf{\#frame} & \textbf{\#pose annotation} & \textbf{\#seq} & \textbf{Object classes}            \\ \hline
\hline

Tom-Net          & \multirow{2}{*}{RGB}   & \multirow{2}{*}{14}        & \multirow{2}{*}{14}  & \multirow{2}{*}{876} & \multirow{2}{*}{seg only} & \multirow{2}{*}{-} & \multirow{2}{*}{Household}         \\
\citep{chen-2018-cvpr}                  &                        &                            &                      &                      &                           &                    &                                    \\ \hline

GDD          & \multirow{2}{*}{RGB}   & \multirow{2}{*}{\ding{55}}        & \multirow{2}{*}{\ding{55}}  & \multirow{2}{*}{3,916} & \multirow{2}{*}{seg only} & \multirow{2}{*}{-} & \multirow{2}{*}{In-the-wild glass}         \\
\citep{mei-2020-cvpr}                  &                        &                            &                      &                      &                           &                    &                                    \\ \hline

Trans10K          & \multirow{2}{*}{RGB}   & \multirow{2}{*}{\ding{55}}        & \multirow{2}{*}{\ding{55}}  & \multirow{2}{*}{$\sim$10K } & \multirow{2}{*}{seg only} & \multirow{2}{*}{-} & Household         \\
\citep{xie-2020-eccv}                 &                        &                            &                      &                      &                           &                    &   In-the-wild glass                                 \\ \hline

GSD          & \multirow{2}{*}{RGB}   & \multirow{2}{*}{\ding{55}}        & \multirow{2}{*}{\ding{55}}  & \multirow{2}{*}{4,012 } & \multirow{2}{*}{seg only} & \multirow{2}{*}{-} & \multirow{2}{*}{In-the-wild glass }         \\
\citep{lin-2021-cvpr}                 &                        &                            &                      &                      &                           &                    &                                    \\ \hline

TACO          & \multirow{2}{*}{RGB}   & \multirow{2}{*}{\ding{55}}        & \multirow{2}{*}{\ding{55}}  & \multirow{2}{*}{1,499} & \multirow{2}{*}{seg only} & \multirow{2}{*}{-} & \multirow{2}{*}{Trash}         \\
\citep{proencca-2020-arxiv}                   &                        &                            &                      &                      &                           &                    &                                    \\ \hline

TransTourch          & \multirow{2}{*}{RGB}   & \multirow{2}{*}{9}        & \multirow{2}{*}{9}  & \multirow{2}{*}{180} & \multirow{2}{*}{seg only} & \multirow{2}{*}{-} & \multirow{2}{*}{Household}         \\
\citep{jiang-2022-tom}                   &                        &                            &                      &                      &                           &                    &                                    \\ \hline

Zerowaste          & \multirow{2}{*}{RGB}   & \multirow{2}{*}{\ding{55} }        & \multirow{2}{*}{\ding{55} }  & \multirow{2}{*}{$\sim$11K} & \multirow{2}{*}{seg only} & \multirow{2}{*}{-} & \multirow{2}{*}{Trash}         \\
\citep{bashkirova-2022-cvpr}                    &                        &                            &                      &                      &                           &                    &                                    \\ \hline

TransCG          & \multirow{2}{*}{RGB-D}   & \multirow{2}{*}{51 }        & \multirow{2}{*}{51}  & \multirow{2}{*}{$\sim$58K} & \multirow{2}{*}{$\sim$0.2M} & \multirow{2}{*}{130} & \multirow{2}{*}{Household}         \\
\citep{fang-2022-ral}                    &                        &                            &                      &                      &                           &                    &                                    \\ \hline

ClearGrasp          & \multirow{2}{*}{RGB-D}   & \multirow{2}{*}{10}        & \multirow{2}{*}{10}  & \multirow{2}{*}{286} & \multirow{2}{*}{736} & \multirow{2}{*}{-} & \multirow{2}{*}{Household, Trash}         \\
\citep{sajjan-2020-icra}                    &                        &                            &                      &                      &                           &                    &                                    \\ \hline

ClearPose          & \multirow{2}{*}{RGB-D}   & \multirow{2}{*}{63}        & \multirow{2}{*}{72}  & \multirow{2}{*}{$\sim$350K} & \multirow{2}{*}{$\sim$5.0M} & \multirow{2}{*}{51} &  Household        \\
\citep{chen-2022-eccv}                    &                        &                            &                      &                      &                           &                    &   Chemical Equipment                                  \\ \hline

TODD          & \multirow{2}{*}{RGB-D}   & \multirow{2}{*}{6}        & \multirow{2}{*}{6}  & \multirow{2}{*}{$\sim$15K} & \multirow{2}{*}{$\sim$0.1M} & \multirow{2}{*}{49} & \multirow{2}{*}{Chemical Equipment}         \\
\citep{xu-2022-corl}                   &                        &                            &                      &                      &                           &                    &                                    \\ \hline

Dex-NeRF          & \multirow{2}{*}{RGB-D}   & \multirow{2}{*}{6}        & \multirow{2}{*}{6}  & \multirow{2}{*}{516} & \multirow{2}{*}{\ding{55}} & \multirow{2}{*}{8} & Household         \\
\citep{ichnowski-2022-corl}                   &                        &                            &                      &                      &                           &                    &       Chemical Equipment                      \\ \hline

TRANS-AFF          & \multirow{2}{*}{RGB-D}   & \multirow{2}{*}{8}        & \multirow{2}{*}{8}  & \multirow{2}{*}{1,346} & \multirow{2}{*}{seg only} & \multirow{2}{*}{-} & \multirow{2}{*}{Household}         \\
\citep{jiang-2022-ral}                  &                        &                            &                      &                      &                           &                    &                                    \\ \hline

STD          & \multirow{2}{*}{RGB-D}   & \multirow{2}{*}{22}        & \multirow{2}{*}{50}  & \multirow{2}{*}{$\sim$27K} & \multirow{2}{*}{$\sim$139K} & \multirow{2}{*}{30} & \multirow{2}{*}{Household}         \\
\citep{dai-2022-eccv}                  &                        &                            &                      &                      &                           &                    &                                    \\ \hline

StereoObj1M          & \multirow{2}{*}{Stereo RGB}   & \multirow{2}{*}{7}        & \multirow{2}{*}{18}  & \multirow{2}{*}{$\sim$393K} & \multirow{2}{*}{$\sim$1.5M} & \multirow{2}{*}{182} & \multirow{2}{*}{Chemical Equipment}         \\
\citep{liu-2021-iccv}                 &                        &                            &                      &                      &                           &                    &                                    \\ \hline

TOD          & \multirow{2}{*}{Stereo RGB-D}   & \multirow{2}{*}{15}        & \multirow{2}{*}{15}  & \multirow{2}{*}{$\sim$48K} & \multirow{2}{*}{$\sim$0.1M} & \multirow{2}{*}{40} & \multirow{2}{*}{Household}         \\
\citep{liu-2020-cvpr}                 &                        &                            &                      &                      &                           &                    &                                    \\ \hline
Polarised          & RGB   & \multirow{2}{*}{6}        & \multirow{2}{*}{6}  & \multirow{2}{*}{1,600} & \multirow{2}{*}{seg only} & \multirow{2}{*}{-} & \multirow{2}{*}{Household}         \\
\citep{kalra-2020-cvpr}                 &     Polarisation                   &                            &                      &                      &                           &                    &                                    \\ \hline

RGB-P          & RGB   & \multirow{2}{*}{\ding{55}}        & \multirow{2}{*}{\ding{55}}  & \multirow{2}{*}{4,511} & \multirow{2}{*}{seg only} & \multirow{2}{*}{-} & \multirow{2}{*}{In-the-wild glass}         \\
\citep{mei-2022-cvpr}                 &     Polarisation                   &                            &                      &                      &                           &                    &                                    \\ \hline

PhoCaL          & RGB-D   & \multirow{2}{*}{8}        & \multirow{2}{*}{60}  & \multirow{2}{*}{3,951} & \multirow{2}{*}{$\sim$91K} & \multirow{2}{*}{20} & \multirow{2}{*}{Household}         \\
\citep{wang-2022-cvpr}                 &     Polarisation                   &                            &                      &                      &                           &                    &                                    \\ \hline

TransCut         & \multirow{2}{*}{Light-field}   & \multirow{2}{*}{7}        & \multirow{2}{*}{7}  & \multirow{2}{*}{49} & \multirow{2}{*}{seg only} & \multirow{2}{*}{-} & \multirow{2}{*}{Household}         \\
\citep{xu-2015-iccv}                &                        &                            &                      &                      &                           &                    &                                    \\ \hline

ProLIT         & \multirow{2}{*}{Light-field}   & \multirow{2}{*}{6}        & \multirow{2}{*}{6}  & \multirow{2}{*}{300} & \multirow{2}{*}{421} & \multirow{2}{*}{-} & \multirow{2}{*}{Household}         \\
\citep{zhou-2020-ral}               &                        &                            &                      &                      &                           &                    &                                    \\ \hline
RGB-T          & RGB   & \multirow{2}{*}{\ding{55}}        & \multirow{2}{*}{\ding{55}}  & \multirow{2}{*}{5,551} & \multirow{2}{*}{seg only} & \multirow{2}{*}{-} & \multirow{2}{*}{In-the-wild glass}         \\
\citep{huo-2023-tip}                 &     \ac{TIR}                   &                            &                      &                      &                           &                    &                                    \\ \hline

\textbf{TRansPose}          & \textbf{Stereo RGB-D}   & \multirow{2}{*}{\textbf{99}}        & \multirow{2}{*}{\textbf{111}}  & \multirow{2}{*}{\textbf{$\sim$100K}} & \multirow{2}{*}{\textbf{$\sim$3.9M}} & \multirow{2}{*}{\textbf{87}} &  \textbf{Household, Trash}         \\
\textbf{(ours)}                &     \textbf{\ac{TIR}}                   &                            &                      &                      &                           &                    & \textbf{Chemical Equipment}                     \\ \hline

\end{tabular}%
}
\end{adjustbox}

\label{tab:comparison}
\end{table*}

\begin{figure}[!t]
  \centering
  \begin{minipage}{0.95\columnwidth} 
  \centering
  \subfigure[Overall System Configuration]
  {%
  \includegraphics[width=0.495\columnwidth]{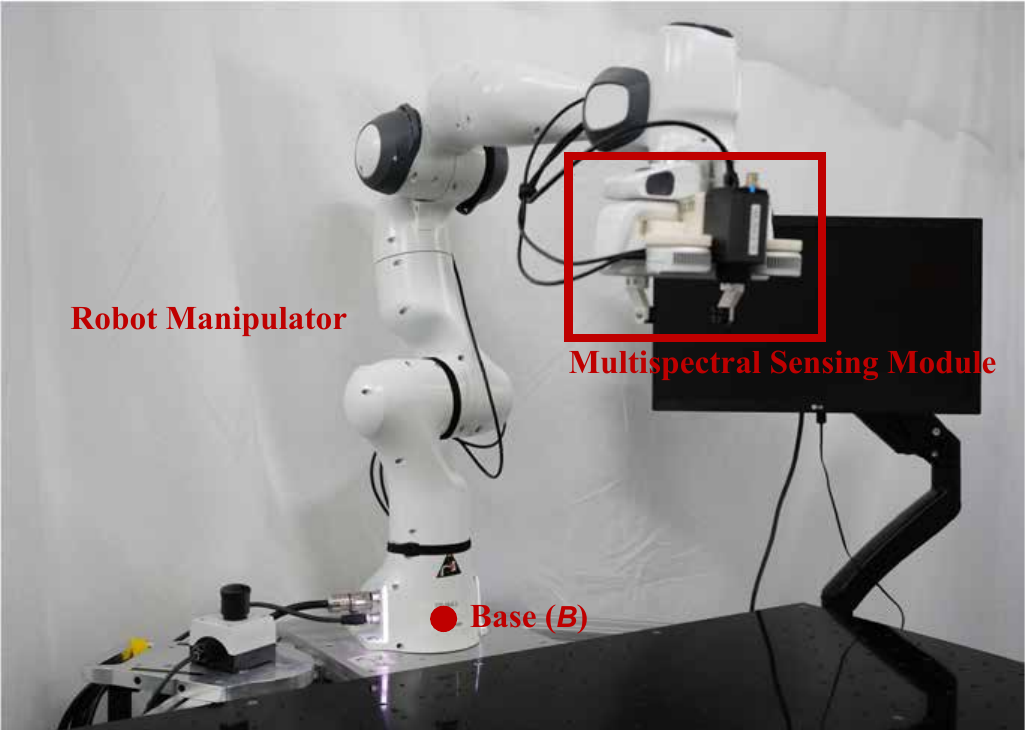}
  \label{fig:system}
  }
  \subfigure[Multispetral Sensing System]
  {%
  \includegraphics[width=0.46\columnwidth]{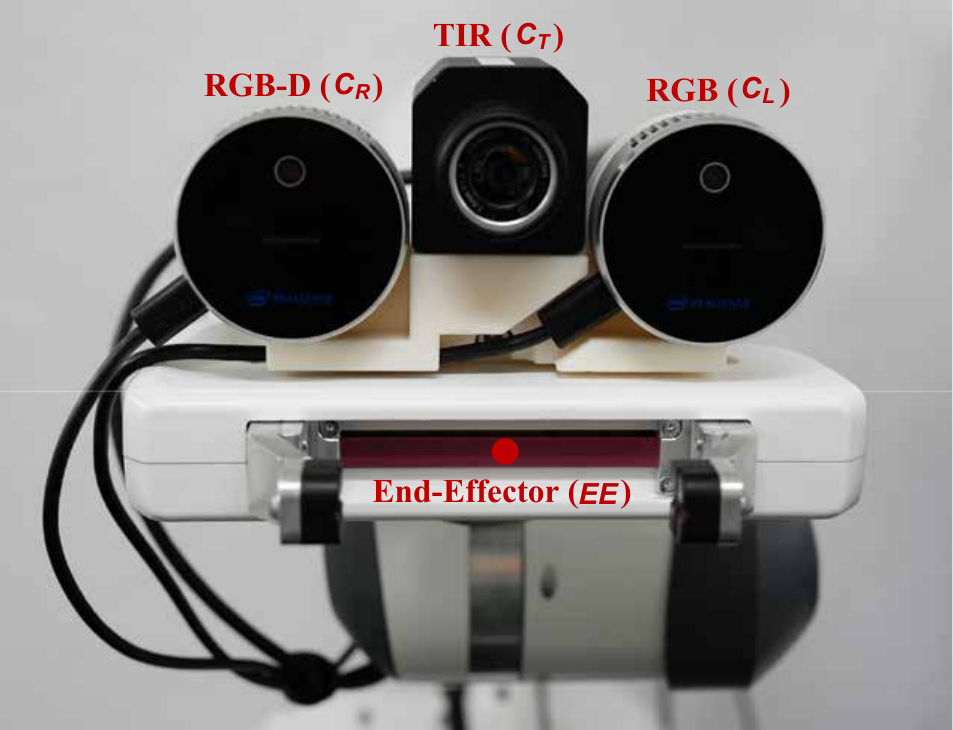}
  \label{fig:sensor}
  }
  \end{minipage}
  \caption{System configuration for the TRansPose dataset. A coordinate system is defined based on the base of the robot manipulator.}
  \label{fig:overall_system}
\end{figure}

In our work, we propose a large-scale multispectral dataset for transparent object recognition, TRansPose (\B{T}IR-\B{R}GB dataset for Tr\B{ans}parent object \B{Pose}).
We exploit two RGB-D cameras and one \ac{TIR} camera attached to the end-effector of the robot manipulator (\figref{fig:overall_system}). Using this multispectral sensing system, we collected data sequences with accurate camera poses relative to the base of the robot manipulator. The proposed dataset includes 99 transparent objects of three types; 43 household objects, 27 recyclable trashes, and 29 chemical laboratory equipment. We provide 87 diverse data sequences with required annotations, including object class, segmentation, and 6D pose. We also offer rendering code to generate result plots as \figref{fig:result}. Furthermore, our dataset encompasses typical yet highly demanding scenarios involving transparent objects enclosed within rigid or deformable containers (e.g., objects in a container, objects in a plastic bag, and multi- stacked objects). The proposed dataset aims to push the boundaries of transparent object recognition research and facilitate advancements in the field of robotics, opening up new possibilities for the perception and manipulation of transparent objects. Compared with existing datasets, the proposed dataset possesses the following key contributions:

\begin{enumerate}
\item We introduce a large-scale multispectral transparent object dataset, TRansPose, incorporating two RGB-D cameras and one \ac{TIR} camera. This camera configuration makes it straightforward to exploit the object shape information.  
\item We provide 99 transparent objects of three types (43 household objects, 27 recyclable trashes, and 29 chemical laboratory equipment) and 12 non-transparent objects.
\item We provide 87 data sequences on the table-top setup for practical robot manipulation tasks. These sequences include various challenging environments; diverse lighting conditions, heavy clutters, objects in the container, objects in plastic-bag, object filled with water and multi-stacked objects.
\item For each data sequence, we provide multispectral images, instance-level segmentation, ground-truth pose, completed depth, as well as 3D object models.
\end{enumerate}

\section{Related Works}

In recent years, datasets for transparent object recognition have proliferated. \tabref{tab:comparison} summarizes recent datasets for transparent object recognition. As mainstream, most researchers were focused on the RGB-based datasets \citep{chen-2018-cvpr, mei-2020-cvpr, liu-2020-cvpr, xie-2020-eccv, sajjan-2020-icra, proencca-2020-arxiv, lin-2021-cvpr, liu-2021-iccv, jiang-2022-elsevier, bashkirova-2022-cvpr, fang-2022-ral, chen-2022-eccv, xu-2022-corl, jiang-2022-ral, dai-2022-eccv, chen-2022-eccv}. These datasets have necessitated background clutters using additional artifacts to overcome the attribute of the transparent objects heavily dependent on the background. Notably, ClearPose \citep{chen-2022-eccv} made significant strides by providing a comprehensive collection of 62 transparent objects, including those commonly encountered daily and used in controlled environments. This dataset not only offers precise pose annotations but also includes instance segmentation. What sets ClearPose apart is its inclusion of diverse objects, encompassing variations in size, shape, and category. However, these RGB-based datasets do not adequately address the challenges arising from the unpredictable surface characteristics of transparent objects and their susceptibility to disruptions caused by the background.

Some studies have also incorporated alternative sensor modalities to tackle challenges provoked by light reflection and refraction in transparent objects. Notably, light-field cameras capable of capturing both the intensity and direction of light have proven valuable in capturing the position and shape of transparent objects \citep{xu-2015-iccv, zhou-2020-ral}. Similarly, polarized cameras have emerged as a suitable choice for overcoming issues related to light reflection and glare while simultaneously enhancing contrast \citep{kalra-2020-cvpr, mei-2022-cvpr, wang-2022-cvpr}. However, the nature of all visible light, such as reflections and refractions within transparent objects and susceptibility to background interference, invoke challenges when applying datasets to transparent object perception tasks, resulting in noise-contaminated and incomplete outcomes.

To address these limitations, \citet{huo-2023-tip} proposed RGB-T dataset which exploits RGB and \ac{TIR} camera pair. Utilizing the inherent characteristic of \ac{TIR} cameras with their zero transmittance for transparent objects, this dataset is harnessed to detect transparent entities, but only for window glasses.

Most of the existing datasets are restricted to specific domains due to the deficient number of objects. These also suffer from a lack of realism, artificially arranged and labeled objects and scenes deviating from real-world conditions. Since these datasets fail to capture the full spectrum of transparent objects found in everyday scenarios, it is arduous to reflect the real-world scenarios commonly encountered daily, such as occlusions, diverse lighting conditions, and complex object arrangements. In our work, we provide 99 transparent objects considering more categories than other existing datasets and challenging scenes, including objects filled with water, diverse lighting conditions, heavy clutters, objects in non-transparent or translucent containers, objects in plastic-bags, and multi-stacked objects. Additionally, our work provides about 3.9M pose annotations for target objects presented in all sequences.

\section{System Description}

\subsection{System Setup}

\figref{fig:overall_system} shows an overall system configuration. We incorporate two Intel RealSense L515 RGB cameras and a FLIR A65 \ac{TIR} camera as a multispectral sensing module. To avoid laser interference between these two sensors, only one camera $C_R$ obtains depth maps. The robot manipulator is installed on the workspace table, which is 1250 $\times$ \unit{850}{mm}. The multispectral sensing module is mounted on the end-effector of the robot manipulator. The sensor specifications used in the proposed dataset are summarized in \tabref{tab:sensor_spec}.

\begin{table}[!t]
    \centering
    \caption{Sensors used in the multispectral sensing module. }
    \begin{adjustbox}{width=1\linewidth}
    {
        \begin{tabular}{cc|cccccc}
        \hline
        \multicolumn{2}{c|}{Sensors}                          & Type               & Manufacturer    & Model & FPS    & Resolution       & Data used   \\ \hline \hline
        \multicolumn{1}{c|}{\multirow{3}{*}{Camera}} & $C_R$  & RGB-D              & Intel Realsense & L515  & 30Hz   & 640 $\times$ 480 & RGB, Depth  \\
        \multicolumn{1}{c|}{}                        & $C_T$ & TIR                & FLIR            & A65   & 30Hz   & 640 $\times$ 512 & 14bit TIR   \\
        \multicolumn{1}{c|}{}                        & $C_L$   & RGB-D              & Intel Realsense & L515  & 30Hz   & 640 $\times$ 480 & RGB         \\ \hline
        \multicolumn{2}{c|}{Robot Manipulator}                & Positional Encoder & Franka Emika    & Panda & 1000hz & 14bit            & Joint Angle \\ \hline
        \end{tabular}
    }
    \end{adjustbox}
    
    \label{tab:sensor_spec}
\end{table}

\subsection{System Calibration}

To utilize the image set obtained from the multispectral sensing module, we need to estimate camera parameters and the extrinsic relationship between all sensors. 

\textbf{Intrinsic Calibration:} For camera parameters of RGB-D cameras, camera calibration is performed using a checkerboard \citep{zhang-1999-iccv}. For the \ac{TIR} camera, we encounter a challenge as it cannot accurately detect the corners of a conventional checkerboard. To surmount this limitation, we utilize a manually designed checkerboard composed of different materials \citep{saponaro-2015-icip}, each capable of undergoing selective heating. The average re-projection calculated with the estimated camera parameters is approximately 0.2 pixels.

\textbf{Extrinsic Calibration:} We execute the hand-eye calibration \citep{tsai-1989-tro} for the extrinsic calibration between all sensors, obtaining three transformations ($^{EE}_{C_R}T, ^{EE}_{C_T}T$ and $^{EE}_{C_L}T$). We fix a calibration pattern on the workspace table and manually capture the images by moving the robot manipulator. For the RGB-D cameras, we use a ChArUco board. For the \ac{TIR} camera, we use the same checkboard with the process of intrinsic calibration. The average rotation and translation error are approximately $0.3^{\circ}$ and \unit{2}{mm}, respectively.

\begin{figure*}[!t]
  \centering
  \begin{adjustbox}{width=0.95\textwidth}
  {
      \begin{minipage}{0.4\textwidth} 
      \centering
      \subfigure[Household Objects]
      {%
          \includegraphics[width=0.47\columnwidth]{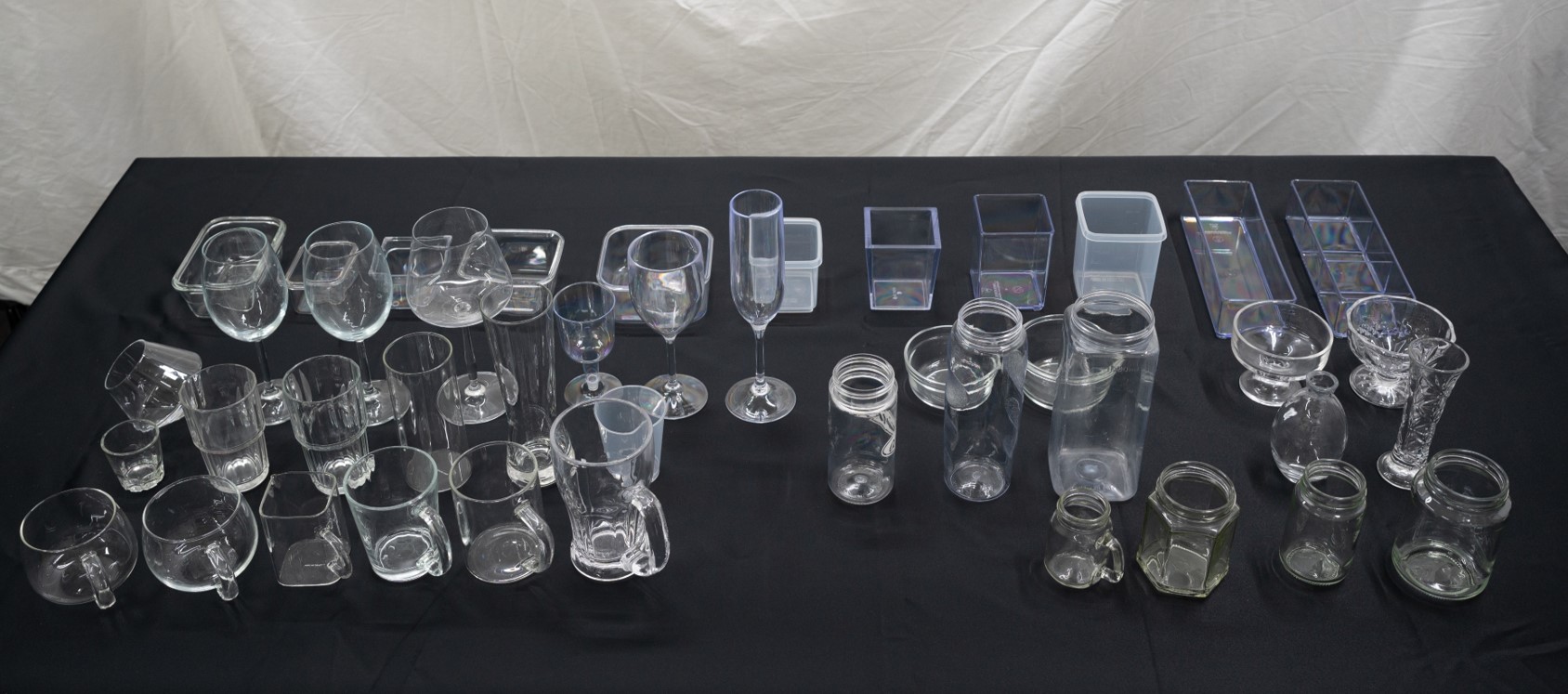}
          \label{fig:h_obj}
      }
      \subfigure[Chemical Lab Equipment]
      {%
          \includegraphics[width=0.47\columnwidth]{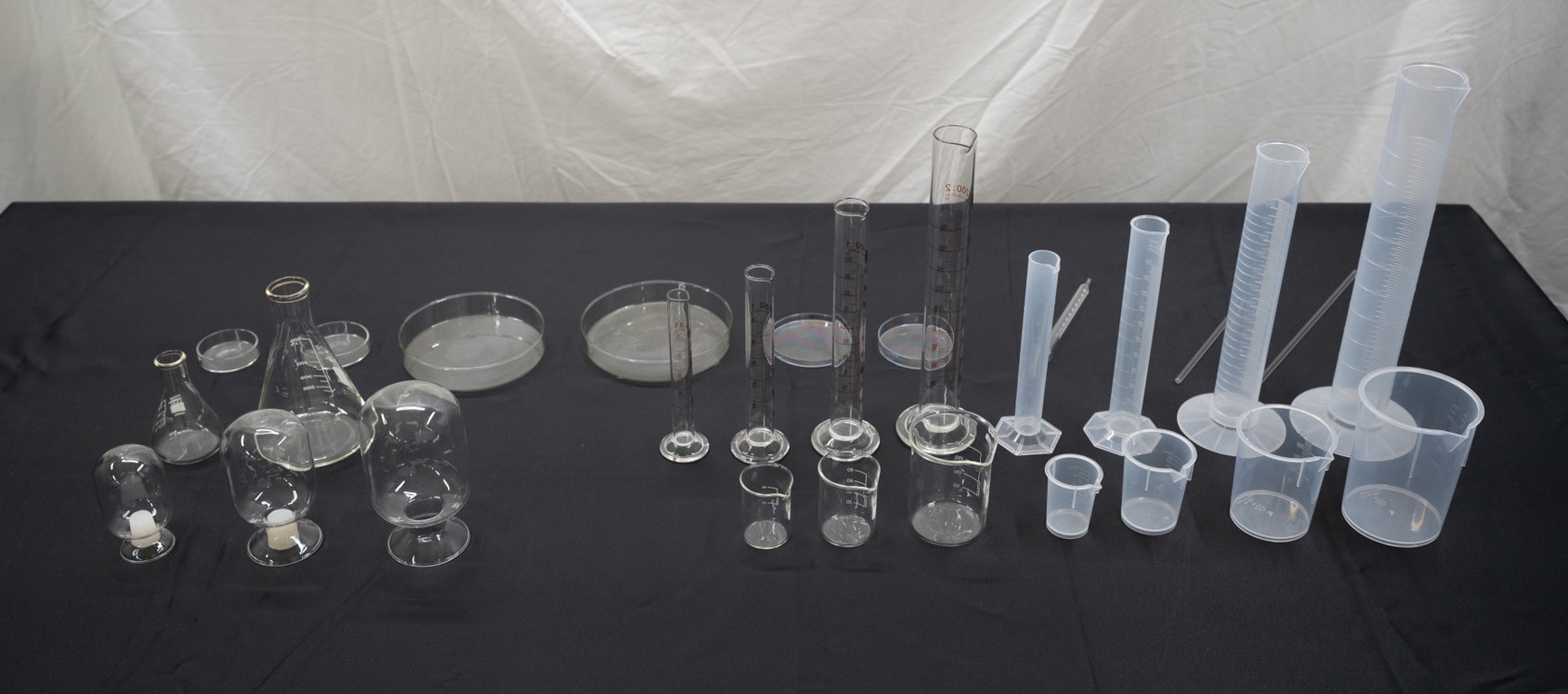}
          \label{fig:c_obj}
      }
      \subfigure[Recyclable Trashes]
      {%
          \includegraphics[width=0.47\columnwidth]{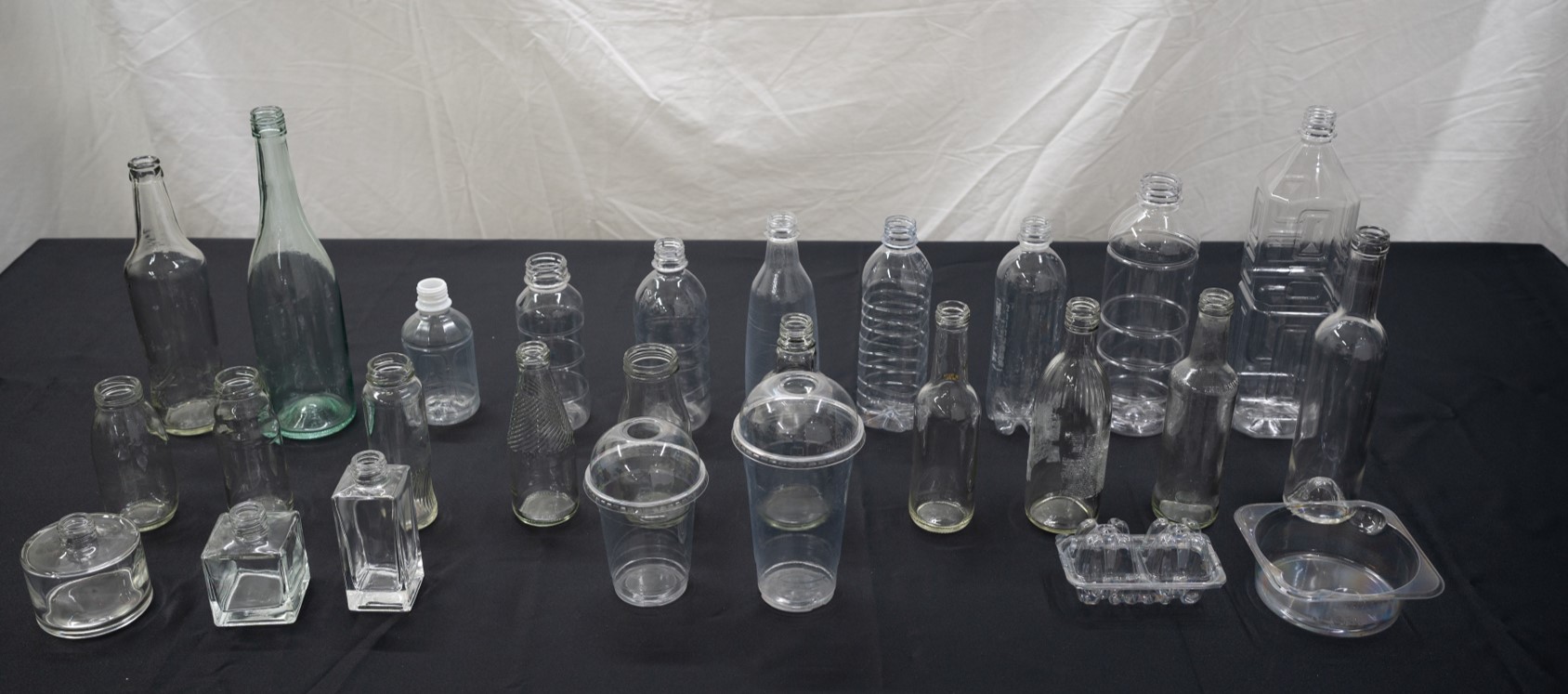}
          \label{fig:t_obj}
      }
      \subfigure[Non-Transparent Objects]
      {%
          \includegraphics[width=0.47\columnwidth]{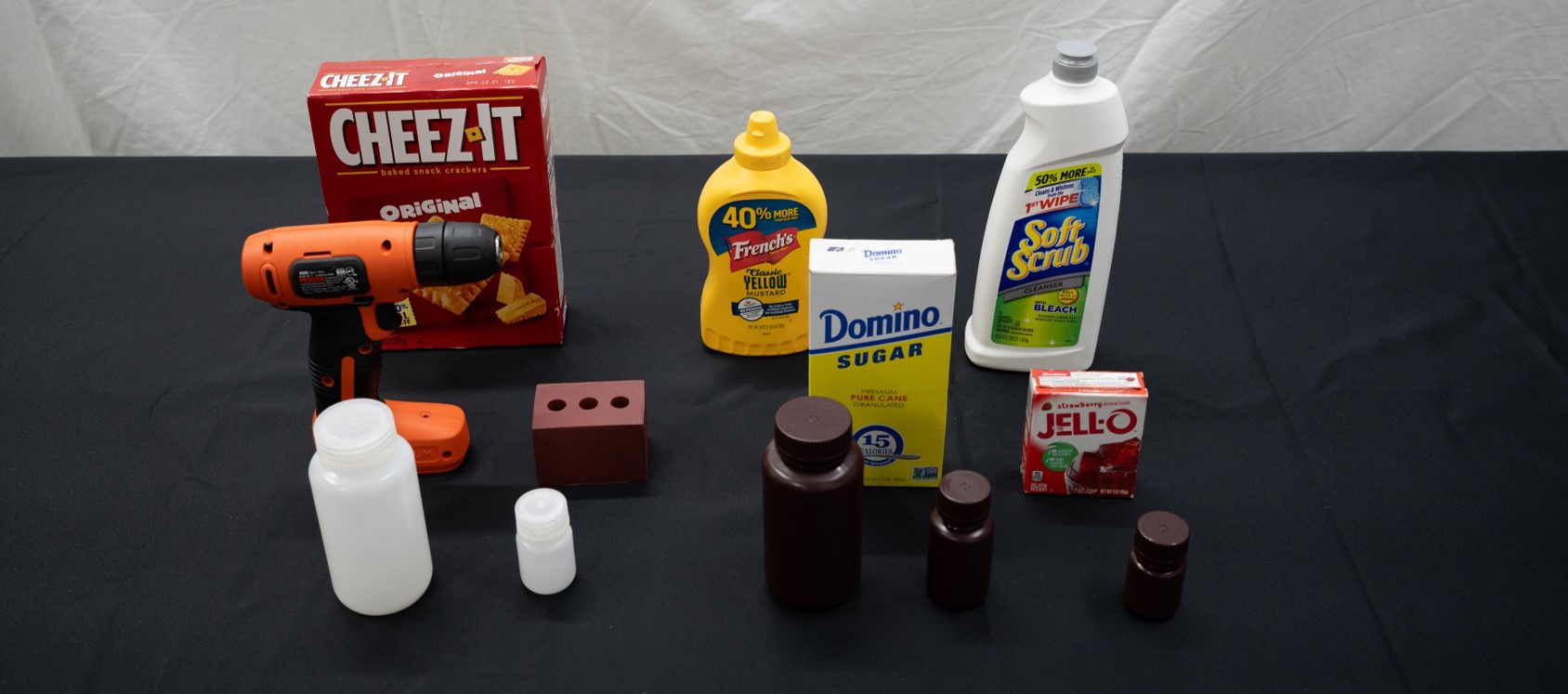}
          \label{fig:n_obj}
      }
      \end{minipage}
  
      \hfill
      
      \begin{minipage}{0.6\textwidth} 
      \centering
      {%
          \subfigure[Pose Annotations]
          {%
          \includegraphics[width=1\columnwidth]{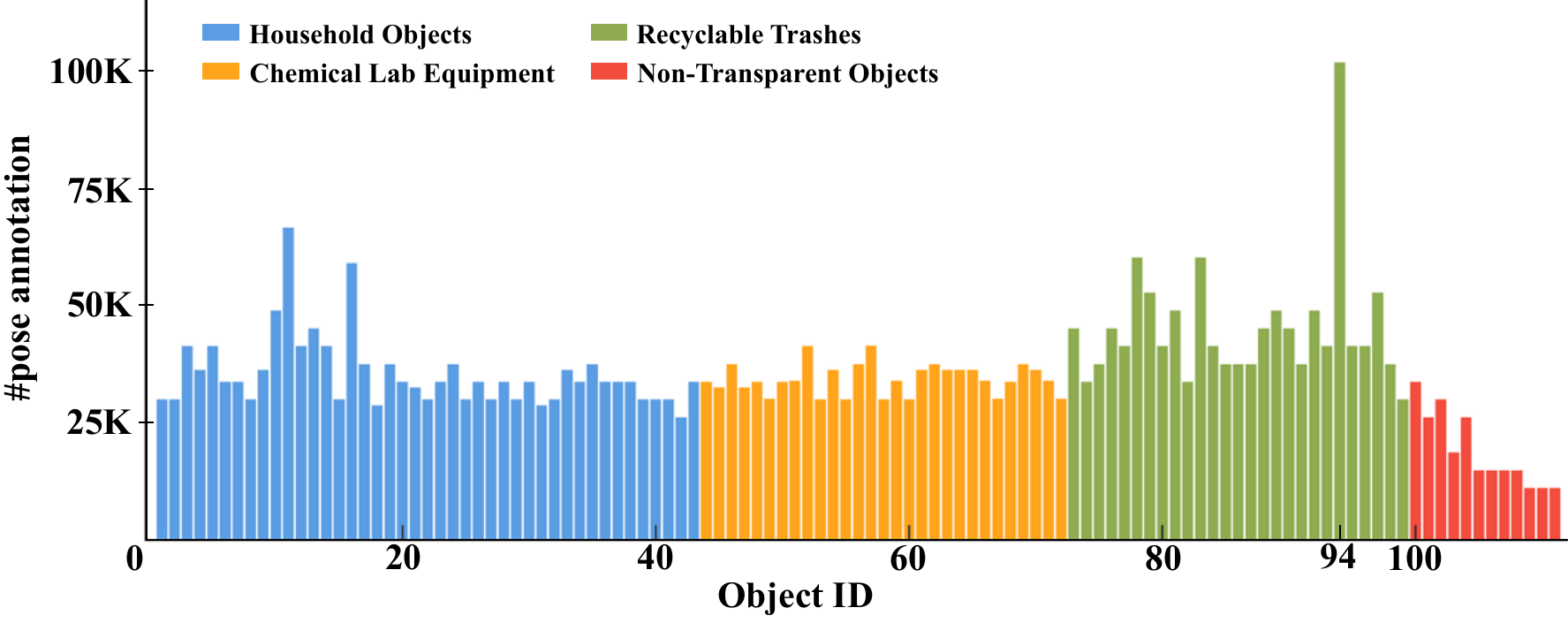}
          \label{fig:distribution}
          }
      }
      \end{minipage}
  }
  \end{adjustbox}

  \caption{(a) - (d) The objects included in the TRansPose dataset. (e) The number of pose annotations per object. In some sequecnes, as multiple-instance of the object \texttt{94} appears, its pose annotations are numerous.}
  \label{fig:object}  
\end{figure*}

\subsection{Data Collection}

3D CAD models of the objects are generated using a 3D scanner. To scan the transparent objects, we cover their surface with a suitable material to ensure complete mesh generation. Additionally, for objects like wine glasses or cylinders where scanning the interior is difficult, we fill them out with additional materials to create a solid mesh representation.

For sequential data acquisition, synchronization between sensors is crucial for the proper usage of image set. However, we confront two problems that prevent the synchronization of all sensors. First, the framerate of each sensor is different, so synchronization is required through external signals. Unfortunately, the multispectral sensing module composed in this work does not inherently support this. Second, the \ac{TIR} camera is susceptible to noise caused by internal heat, resulting in increased image noise over time. To address this issue, this sensor incorporates \ac{NUC} function to compensate for the inherent noise and improve image quality. This process causes a delay of approximately \unit{1}{\s} in the data acquisition, making it challenging to synchronize with other sensors. 

To overcome the above problems, our work involves discrete data capturing. In advance, we pre-define 14 waypoints as the trajectory of the robot manipulator. We set up 90 steps between each waypoint, capturing data at every step. As a result, joint angles of the robot manipulator, two RGB images, a depth image, and a \ac{TIR} image can be obtained simultaneously at \unit{30}{fps} without motion blur. 

\section{Data Annotation}

Most target objects have transparent material types, so we cannot obtain proper depth maps, making it challenging to align the 6D object poses. To tackle this, we employ ProgressLabeller to annotate the 6D object poses ($^{C_R}_{obj}T$, $^{C_T}_{obj}T$ and $^{C_L}_{obj}T$), instance-segmentation masks, and ground truth depth maps. ProgressLabeller is a manual labeling tool for 6D object pose plugged into Blender, proposing a sophisticated multi-view silhouette matching technique to align objects in a 3D interactive workspace. Since ProgressLabeller utilizes the monocular RGB-D sequence only for object pose labeling, we enhance the ProgressLabeller to utilize stereo RGB and \ac{TIR} images and the end-effector pose of the robot manipulator simultaneously. We import the following: stereo RGB and \ac{TIR} images, 3D CAD models, camera poses, and intrinsic and extrinsic parameters. Since ProgressLabeller utilizes visual \ac{SLAM} to estimate the camera poses, it cannot always ensure to estimate accurate camera poses. In our work, we utilize the end-effector pose of the robot manipulator and extrinsic transformations ($^{EE}_{C_R}T, ^{EE}_{C_T}T$ and $^{EE}_{C_L}T$) between the end-effector of the robot manipulator and three cameras. As a result, we can produce more accurate and consistent annotations.

Manually aligning the projected object model across all multi-view images is time-consuming and requires a lot of delicacy to make accurate annotations. To alleviate these issues, we implement object-wise mask-based optimization. For this process, we manually annotate object mask for approximately five to eight images captured from various viewpoints, encompassing both TIR and stereo RGB. This process is executed in only each single image so that it is more easier than the aligning process across all multi-view images. Then, we optimize the 6D object pose, minimizing the loss $\mathcal{L}$ between the images obtained by projecting the CAD model of the target object and the labeled mask. 

We define a rendering operator as $\mathcal{S} = Render(K, T, P)$, which render object points $P$ given camera intrinsic $K$ and camera pose $T$ into an object mask $\mathcal{S}$. Assuming that ${}_{X}^{EE}{T}$ and ${}_{EE}^{World}{T}$ are already known and $_{Obj}^{X}{T}_{i}$ is obtained by manual labeling process, ${}_{Obj}^{World}{\overline{T}}$ is estimated by minimizing below loss.

\small
\begin{eqnarray}
&\mathcal{L} = \sum_{i,X}||Render(K_X, {}_{X}^{EE}{T}^{-1} {}_{EE}^{World}{T}^{-1} {}_{Obj}^{World}{\overline{T}}, P_{Obj}) \nonumber \\ & - Render(K_X, {}_{Obj}^{X}{T}_{i}, P_{Obj}) ||^2
\label{loss_function}
\end{eqnarray}
\normalsize

, $P_{Obj}$ and $K_X$ represent a set of object points and camera intrinsic parameters respectively. $X$ represents the type of camera, with $X \in \{C_L, C_T, C_R\}$.

Object-wise mask-based optimization proves to be more effective in challenging scenes such as multi-stack or container scenes in terms of accuracy and time consumption, while manual labeling is only effective in simpler scenes.

Based on the labeled 6D object pose, we produce all annotations for object-wise instance segmentation masks and completed depth map. The completed depth maps are generated by replacing the object parts of raw depth images with rendered depth images using object CAD models.

\begin{figure*}[!t]
  \centering
  \begin{minipage}{0.95\textwidth} 
  \centering
  \subfigure[]
  {%
  \includegraphics[width=0.3\textwidth]{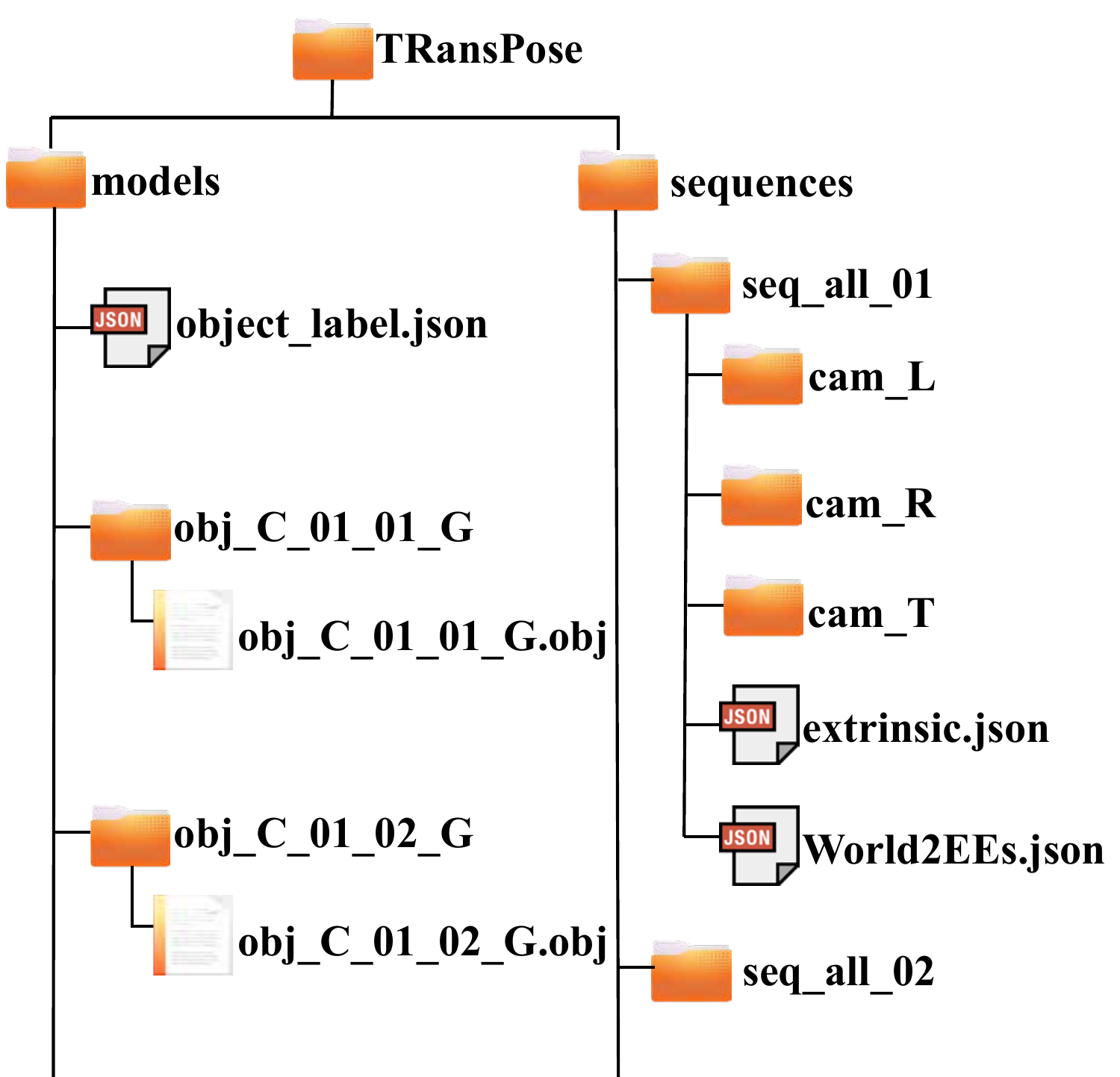}
  \label{fig:TRansPose_annotation}
  }
  \subfigure[]
  {%
  \includegraphics[width=0.35\textwidth]{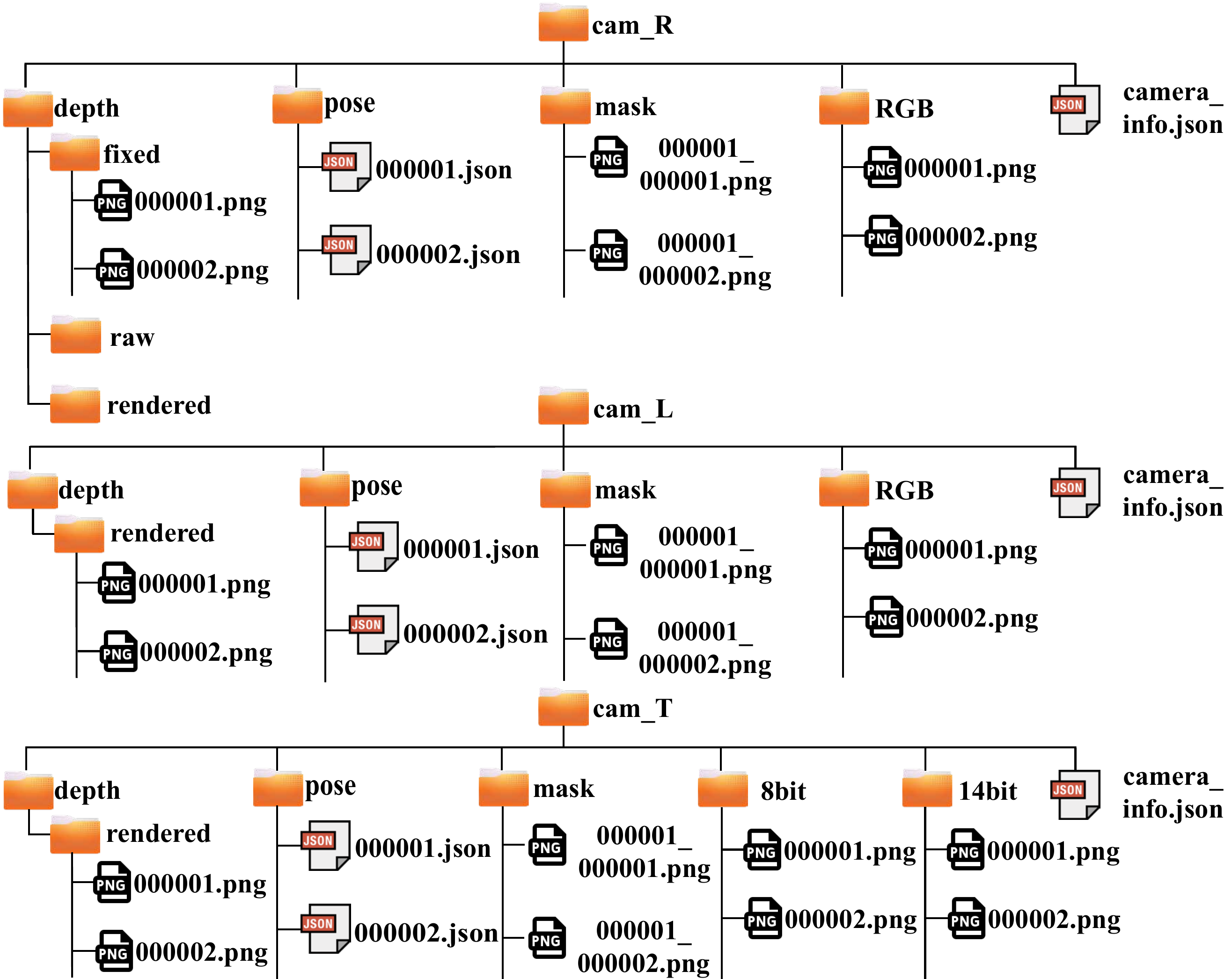}
  \label{fig:cam_annotation}
  }
  \subfigure[]
  {%
  \includegraphics[width=0.27\textwidth]{Figure/Pose_annotation.pdf}
  \label{fig:Pose_annotation}
  }
  \end{minipage}
  \caption{Structure of provided dataset. (a) \texttt{TRansPose} include \texttt{models} and \texttt{sequences} folder. \texttt{models} folder includes 3D CAD models and \texttt{object\_label.json} which has a dictionary paired with the ID and name of objects. In \texttt{sequences} folder, all data is provided according to the sequence index. (b) All data is separated depending on the sensor. For all sensors, images, masks, depth maps, poses, and camera information are provided. Additionally, For the right RGB camera, raw and completed depth maps are provided. In the case of \ac{TIR} camera, we provide 14bit raw images and 8bit images applied simple min-max normalization. (c) Pose annotations are provided as \texttt{json} extension.}
  \label{fig:data_format}
\end{figure*}


\section{Dataset Description}

As shown in \figref{fig:object}, TRansPose dataset consists of 99 transparent objects and 12 non-transparent objects. All objects are further categorized as follows:

\begin{itemize}[leftmargin=2mm, noitemsep]
\item \textbf{43 Household objects} (\figref{fig:h_obj}): 2 vases, 2 dishes, 2 bowls, 3 water bottles, 4 jars, 6 cups with handles, 6 cups without handles, 7 wine glasses, and 11 containers.
\item \textbf{27 Recyclable Trashes} (\figref{fig:t_obj}): 2 disposable cups, 2 pet bottles, 3 perfume containers, and 20 beverage bottles.
\item \textbf{29 Chemical Laboratory Equipment} (\figref{fig:c_obj}): 1 pipette, 2 Erlenmeyer flasks, 2 glass stirring rods, 3 seeds bottles, 6 petri dishes, 7 beakers, and 8 cylinders.
\item \textbf{12 Non-Transparent Objects} (\figref{fig:n_obj}): 5 bottles and 7 general objects used in YCB-Object \citep{calli-2017-ycb}.
\end{itemize}

We provide 87 sequences (333,819 images), which are categorized into four groups: 11 scenes with household objects, 11 scenes with chemical laboratory equipment, 15 scenes with recyclable objects, and 50 scenes with all categories of objects. Additionally, we provide 20 different backgrounds, each characterized by various colors, patterns, and materials such as silk and denim. These sequences encompass several challenging scenes; objects filled with water, diverse lighting conditions, heavy clutters, objects in non-transparent or translucent containers, objects in plastic-bags, and multi-stacked objects. 87 sequences are divided into 61 sequences as trainset and 26 sequences as testset.

\figref{fig:data_format} depicts the structure of the provided data. All data are provided in a compressed file with the \texttt{tar.gz} extension. The 3D CAD models and model ID of all provided objects are located in \texttt{models} folder. All sensor data, related annotations, and calibration values for each sensor are stored in \texttt{sequences} folder. The provided annotations include 6D object poses, instance segmentation masks, bounding-box, and completed depth maps. The distribution of the 6D object pose annotations is shown in \figref{fig:distribution}. Each transparent object in the provided dataset has at least 26K object pose annotations.


\section{Conclusion}

In this paper, we have presented TRansPose, the first large-scale multispectral transparent object dataset, encompassing \ac{TIR} images, stereo RGB-D images, annotated pose, mask, and associated labels. It surpasses existing datasets by offering a wider range of categories and objects, enabling its applicability in various environments, including chemical laboratories, households, and recycling centers. Additionally, TRansPose incorporates challenging scenes featuring diverse lighting conditions, heavy clutter, objects inside translucent or non-transparent containers, objects inside plastic bags, objects filled with water, and multi-stacked objects. This comprehensive dataset aims to facilitate research and advancements in transparent object recognition across a wide array of real-world scenarios.

\begin{acks}
This work was supported by Institute of Information \& communications Technology Planning \& Evaluation (IITP) grant funded by the Korea government(MSIT) No.2022-0-00480, Development of Training and Inference Methods for Goal-Oriented Artificial Intelligence Agents. Also, this work was supported by the National Research Foundation of Korea(NRF) grant funded by the Korea government(MSIP) (2020R1C1C1006620)
\end{acks}

\bibliographystyle{SageH}
\bibliography{string-short,reference}

\end{document}